\documentclass[final]{cvpr}

\usepackage{times}
\usepackage{epsfig}
\usepackage{graphicx}
\usepackage{amsmath}
\usepackage{amssymb}

\usepackage{cite}
\usepackage{url}
\usepackage{bm}
\usepackage{array}
\usepackage{makecell}
\usepackage{booktabs}
\usepackage{multirow}
\usepackage{algorithm}
\usepackage{algorithmic}
\usepackage{color}

\usepackage[pagebackref=true,breaklinks=true,colorlinks,bookmarks=false]{hyperref}

\DeclareMathOperator\dif{d\!}
\newcommand{\PreserveBackslash}[1]{\let\temp=\\#1\let\\=\temp}
\newcolumntype{C}[1]{>{\PreserveBackslash\centering}p{#1}}
\newcolumntype{R}[1]{>{\PreserveBackslash\raggedleft}p{#1}}
\newcolumntype{L}[1]{>{\PreserveBackslash\raggedright}p{#1}}
\newenvironment{shrinkeq}[2]
{ \bgroup
  \addtolength\abovedisplayshortskip{#1}
  \addtolength\abovedisplayskip{#1}
  \addtolength\belowdisplayshortskip{#2}
  \addtolength\belowdisplayskip{#2}}
{\egroup\ignorespacesafterend}

\def\cvprPaperID{2429} 
\def\confYear{CVPR 2021}

\begin{document}

\title{Semi-Supervised Video Deraining with Dynamical Rain Generator}
\author{Zongsheng Yue$^{1}$,  Jianwen Xie$^{2}$, Qian Zhao$^{1}$, Deyu Meng$^{1,3,}$\footnotemark[1] \\ 
$^{1}$Xi'an Jiaotong University, Xi'an, China \\
$^{2}$Cognitive Computing Lab, Baidu Research, Bellevue, USA \\
$^{3}$The Macau University of Science and Technology, Macau, China\\
{\tt\small zsyzam@gmail.com}, {\tt\small jianwen@ucla.edu}, 
{\tt \small timmy.zhaoqian@gmail.com}, {\tt\small dymeng@mail.xjtu.edu.cn} \\
\url{https://github.com/zsyOAOA/S2VD}
}


\maketitle
\pagestyle{empty}
\thispagestyle{empty}
\renewcommand\thefootnote{\fnsymbol{footnote}}
\footnotetext[1]{Corresponding author.}
\renewcommand\thefootnote{\arabic{footnote}}
\thispagestyle{empty}

\begin{abstract}
    While deep learning (DL)-based video deraining methods have achieved significant successes in recent years,
    they still have two major drawbacks. 
    Firstly, most of them are insufficient to model the characteristics of rain layers contained in rainy videos.
    In fact, the rain layers exhibit strong visual properties (e.g., direction, scale, and thickness) in spatial dimension and 
    causal properties (e.g., velocity and acceleration) in temporal dimension, and thus can be modeled by the spatial-temporal process in statistics.
    Secondly, current DL-based methods rely heavily on the labeled training data, whose rain layers are synthetic, thus leading to a deviation from real data.
    Such a gap between synthetic and real data sets results in poor performance when applying them to real scenarios.
    To address these issues, this paper proposes a new semi-supervised video deraining method, in which a dynamical rain generator is employed
    to fit the rain layer for the sake of better depicting its intrinsic characteristics. Specifically, the dynamical generator consists of
    one emission model and one transition model to simultaneously encode the spatial appearance and temporal dynamics of rain streaks, respectively, both of which are parameterized by deep neural networks (DNNs). Furthermore, different prior formats are designed for the labeled synthetic and unlabeled real data so as to fully exploit their underlying common knowledge. Last but not least, we design a Monte Carlo-based EM algorithm to learn the model. Extensive experiments are conducted to verify the superiority of the proposed semi-supervised deraining model.
\end{abstract}

\section{Introduction}

Rain is a very common bad weather that exists in many videos. The appearance of rain not only negatively affects
the visual quality of the video, but also seriously deteriorates the performance of subsequent video processing algorithms,
e.g., semantic segmentation\cite{Mehta2018}, object detection\cite{Dalal2005},
and autonomous driving\cite{Chen2016}. Thus, as a necessary video pre-processing step,
video deraining has attracted much attentions from the computer vision community.

As an ill-posed inverse problem raised by Garg and Nayar\cite{Garg2004}, various methods have been proposed to handle the video deraining
task\cite{wang2019survey}. Most of the traditional methods focus on exploiting rational prior knowledge for the background or rain layers so as to
obtain a proper separation between them. For example, low-rankness\cite{Jiang2017,Wei2017,Jiang2019} is widely used to encode
the temporal correlations of background video. As for rain streaks, many visual characteristics, such as photometric appearance\cite{garg2005does},
geometrical features\cite{ren2017video}, chromatic consistency\cite{liu2009pixel}, local structure correlations\cite{chen2013generalized} and
multi-scale convolutional sparse coding\cite{li2018video}, have been explored in the past few years. Different
from these deterministic assumptions for rain streaks, Wei \textit{et al}.\cite{Wei2017} firstly regard them as random variables, and
use Gaussian mixture model (GMM) to fit them. Albeit substantiated to be effective in some ideal scenarios, these traditional methods
are mainly limited by the subjective manually-designed prior knowledge and huge computation burden.

\begin{figure}[tp]
    \centering
    \includegraphics[scale=0.84]{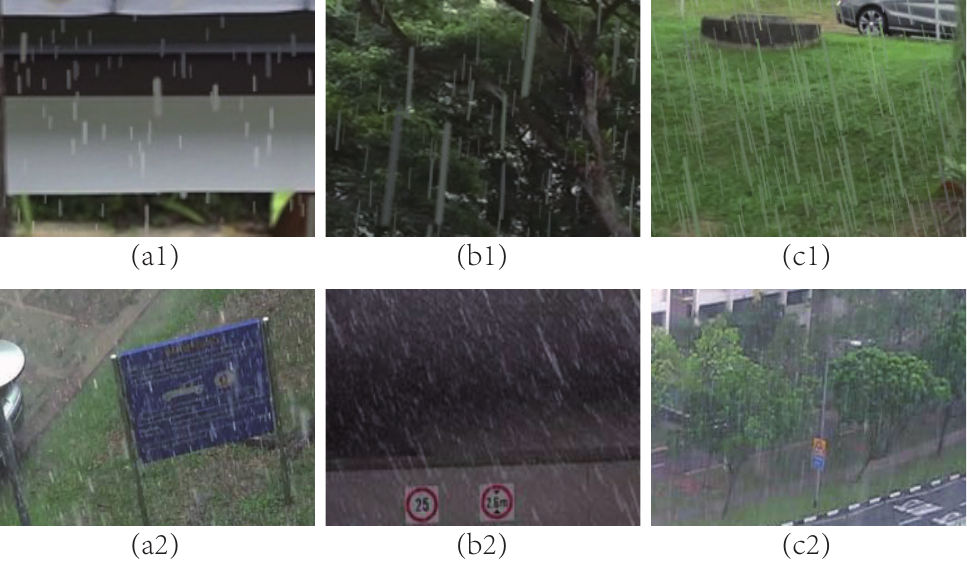}
    \vspace{-1mm}
    \caption{ The comparison of typical synthetic and real rainy images in \textit{NTURain} data set. (a1)-(c1): synthetic rainy images,
    (a2)-(c2): real rainy images.}
    \label{fig:rain_show}
    \vspace{-2mm}
\end{figure}

Recently, owning to the powerful nonlinear fitting capability of DNNs, DL-based methods facilitate significant improvements for the video deraining task.
The core idea of this methodology is to directly train a derainer parameterized by DNNs based on synthetic rainy/clean video pairs in
an end-to-end manner. Most of these methods leverage different technologies, e.g.,
superpixel alignment\cite{chen2018robust}, dual-level flow\cite{yang2019frame} and self-learning\cite{yang2020self},
to extract clean backgrounds from rainy videos. In addition, Liu \textit{et al.}\cite{liu2018erase,liu2018d3r} design a recurrent network to jointly
perform both the rain degradation classification and rain removal tasks.

Even though these DL-based methods have achieved impressive deraining results on some synthetic benchmarks, there still exists large room to further increase the
performance and the generalization capability in real applications. On one hand, most of these methods make efforts to depict the 
background, but neglect to model the intrinsic characteristics of the rain layers.
In fact, the rain layer in video, which is an image sequence of rain steaks, can be represented by a spatial-temporal process. Specifically, the randomly scattered rain streaks in each time frame are characterized with evident visual properties (e.g., direction, scale, and thickness) in the spatial dimension, and the rain layers in different time frames correspond to a continuous time series along the temporal dimension, showing the causal properties (e.g., velocity and acceleration) of the rain dynamics.
Therefore, elaborately representing and exploiting these intrinsic physical properties underlying the rain layers in video data is expected to facilitate the rain removal task.

On the other hand, it is well known that the performance of DL-based methods heavily relies on a large amount of pre-collected training data, i.e.,
rainy/clean video pairs. In fact, due to the high labor cost to obtain such video pairs in real scenes, most of current methods have to use
synthetic ones, which are manually simulated based on the photo-realistic rendering technique~\cite{garg2006photorealistic} or professional
photography and human supervision~\cite{wang2019spatial}. Fig.\ref{fig:rain_show} presents several
typical frames of synthetic and real rainy images in \textit{NTURain}~\cite{chen2018robust} data set, which is widely used as a benchmark
for current video deraining methods. It can be easily seen that the rain patterns in synthetic and real rainy images are obviously different, and the real ones contain more complex and diverse rain types.
Because of such a deviation between synthetic and real data sets, the performances of these DL-based methods deteriorate seriously in the real cases. To deal with the generic video deraining task, it is thus critical to build a reasonable semi-supervised learning framework that
sufficiently exploits the common knowledge in the labeled synthetic and unlabeled real data.

To address these issues, in this paper we propose a semi-supervised video deraining method, in which a dynamical rain generator is
adopted to mimic the generation process of the rain layers in video, hopefully better capturing the intrinsic knowledge simultaneously
from the spatial and temporal dimensions. Besides, the real rainy videos are taken into consideration in our model as unlabeled data, in order to achieve more robust deraining results. In summary, the contributions of this work are as follows:

Firstly, we propose a new probabilistic video deraining method, in which a dynamical rain generator, consisting of a transition model
and an emission model, is employed to fit the rain layers in videos. Specifically, the transition model is used to
represent the dynamics of rains in a low-dimensional state space, while the emission model seeks to generate the observed rain streaks in the image space from the state space. To increase the capacities of such a dynamical rain generator, both the transition and emission models are parameterized by DNNs. Secondly, a semi-supervised learning mechanism is designed by constructing different prior formats for labeled synthetic data and unlabeled real data. Specifically, for the labeled synthetic data, the corresponding ground truth rain-free videos are included  into an elaborate prior distribution as a strong constraint. As for the unlabeled real data, we introduce the 3-D Markov Random Field (MRF) to
model the temporal consistencies and correlations of the underlying backgrounds. Thirdly, a Monte Carlo-based EM algorithm is designed to learn the model. In the expectation step, the posterior of the latent variables is intractable due to the usages of DNNs to parameterize the generator and derainer, thus the Langevin dynamics is adopted to 
approximate the expectation.

\section{Related Work}
In this section, we give a short recap for the developments on the video/image deraining methods.

\subsection{Video Deraining Methods}
To the best of our knowledge, Garg and Nayar\cite{Garg2004} firstly proposed the problem of video deraining, and developed a rain detector based
on the photometric appearance of rain. Later, they further explored the relationships between rain effects and some
camera parameters~\cite{garg2005does,garg2006photorealistic,garg2007vision}.

Inspired by these seminal works, various video deraining methods have been proposed in the past few years, focusing on seeking more reasonable prior knowledge
for the rain or background. For example, both the chromatic properties~\cite{zhang2006rain,liu2009pixel} and shape characteristics~\cite{brewer2008using,bossu2011rain} 
of rain in the time domain were employed to identify and remove the rain layers from the captured rainy 
videos, while the regular visual effects of rain in the global frequency space were also exploited by~\cite{barnum2010analysis}.
Besides, Santhaseelan and Asari~\cite{santhaseelan2015utilizing} employed local phase congruency to detect rain based on chromatic constraints.
Notably, Wei \textit{et al.}\cite{Wei2017} firstly regarded rain streaks as random variables and fitted them by GMM.
In addition, matrix/tensor factorization technologies were also very popular in the field of video deraining, which were typically used to encode the correlations of
background video along the temporal dimension, e.g., ~\cite{chen2013generalized,kim2015video,Jiang2017,Jiang2019,ren2017video}.

In recent years, DL-based methods represent a new trend along this research line. In~\cite{li2018video}, Li \textit{et al.} employed the multi-scale
convolutional sparse coding to encode the repetitive local patterns under different scales of rain streaks. 
Chen \textit{et al.}~\cite{chen2018robust} 
proposed to decompose the scene into superpixels and then align the scene content based on the superpixel segmentation result, and finally a CNN was used to
compensate the lost details and add normal textures to the deraining result. In~\cite{liu2018erase}, Liu \textit{et al.} designed a recurrent
neural network to jointly perform the rain degradation classification and rain removal tasks. And in~\cite{liu2018d3r}, a hybrid rain model
was proposed to model both rain streaks and occlusions. Besides, Yang \textit{et al.}\cite{yang2019frame} also built a two-stage recurrent networks that utlize
dual-level regularizations toward video deraining. Very recently, Yang \textit{et al.}~\cite{yang2020self} proposed a self-learning manner for
this task by taking both temporal correlations and consistencies into consideration.

While DL-based methods have achieved impressive performance on some synthetic benchmarks, they are still very hard to be applied to the real applications due to the large gap between the used synthetic data and the real data. Therefore, in order to increase the generalization capacity of the deraining model in the real tasks, it is crucial to design a semi-supervised learning framework that makes use of the information in both the labeled synthetic data
and the unlabeled real one. This paper mainly focuses on this issue.

\subsection{Single Image Deraining Methods}
For literature comprehensiveness, we also briefly review the single image deraining methods.
The single image deraining method can be roughly divided into two categories, i.e., model-based
methods and DL-based methods. Most of the model-based methods formulate the deraining task as a decomposition problem of the rain and background layers,
and various technologies have been employed to deal with it, such as morphological component analysis~\cite{kang2011automatic}, non-local means
filter~\cite{kim2013single}, and sparse coding~\cite{chen2014visual,luo2015removing}. Besides, methods built on prior knowledge of rain and background are also explored in this field,
mainly including sparsity and
low-rankness~\cite{zhang2017convolutional,chang2017transformed,gu2017joint}, narrow directions of rain and the similarities of rain patches~\cite{zhu2017joint},
and GMM~\cite{li2016rain}.

The earliest DL-based method was proposed by Fu \textit{et al.}~\cite{fu2017clearing,fu2017removing}, in which CNNs were adopted to remove rains from the high
frequency parts of rainy images.
Led by these two works, DL-based methods began to dominate the research in this field. Many effective and advanced
network architectures~\cite{li2018non,li2018recurrent,ren2019progressive,wang2019spatial,fu2019lightweight,Hu_2019_CVPR} were put forward in recent years. And some works attempted to jointly handle the rain removal task with other related tasks, e.g., rain detection~\cite{yang2017deep}, rain density estimation~\cite{zhang2018density},
so as to obtain better deraining performance. Besides, some useful priors, e.g., multi-scale~\cite{Yasarla_2019_CVPR,zheng2019residual,Jiang_2020_CVPR},
convolutional sparse coding~\cite{Wang_2020_CVPR} and bilevel layer prior~\cite{mu2018learning}, were also embedded into the DL-based methods to sufficiently
mine the potentials of DNNs. Different from the above methods, Zhang \textit{et al.}~\cite{2017Image} and Wang \textit{et al.}~\cite{2018Perceptual}
both introduced the adversarial learning scheme to enhance the fidelity of the derained images, and Wei \textit{et al.}~\cite{Wei_2019_CVPR} proposed a semi-supervised
deraining model that can be better generalized to real tasks.

In general, single image deraining methods can be directly used
in the video deraining task by treating each video as a bunch of independent images. However, ignoring the abundant temporal information contained in videos will lead to an unsatisfied performance. Thus it is necessary to design a reasonable deraining model specific for video data.
\begin{figure*}[t]
    \centering
    \includegraphics[scale=0.51]{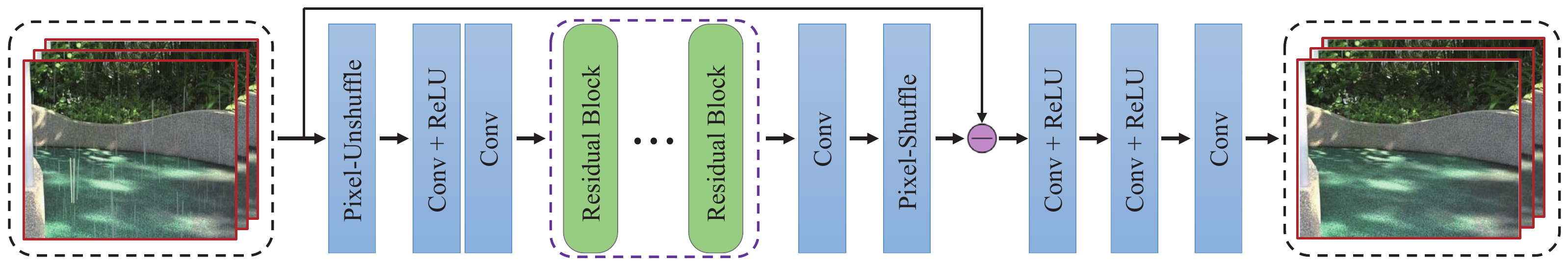}
    \caption{The network architecture for the derainer $f(\cdot;W)$. In this figure, all ``Conv''s denote the 3-D convolution layer.}
    \label{fig:derain_net}
    \vspace{-3mm}
\end{figure*}

\begin{figure*}[t]
    \centering
    \includegraphics[scale=0.57]{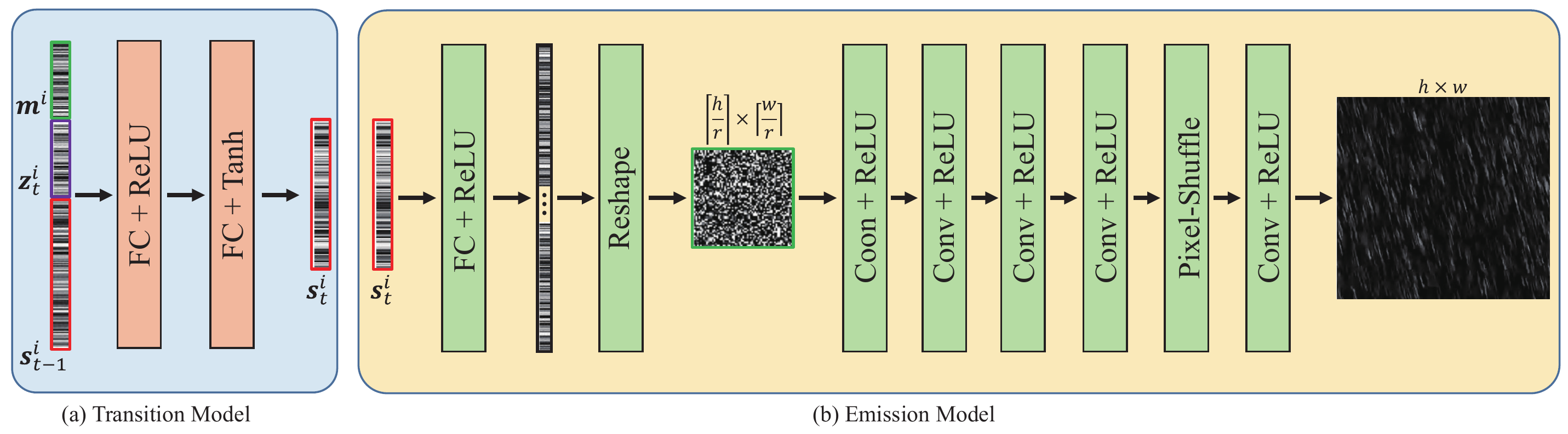}
    \caption{An illustration of network architectures of the transition model and the emission model in the dynamical rain generator. In this figure, ``FC'', ``Conv'' and
    ``Tanh'' denote fully connected, 2-D convolution and hyperbolic tangent layers, respectively. ``Pixel-Shuffle'' is the sub-pixel layer~\cite{shi2016real} with a scale
    factor $r$.}
    \label{fig:generator}
\end{figure*}
\section{Semi-Supervised Video Deraining Model} \label{sec:model}
Given a labeled data set $\mathcal{D}=\{\mathcal{Y}^k, \mathcal{X}^k\}_{k=1}^{N_l}$ and an unlabeled data set $\mathcal{U}=\{\mathcal{Y}^k\}_{k=1}^{N_u}$, where
$\mathcal{Y}^k$ and $\mathcal{X}^k$ denote the $k$-th rainy and clean videos, respectively, we aim to construct a semi-supervised probabilistic
model based on them and then design an EM algorithm to learn the model.

\subsection{Model Formulation}\label{subsec:model}
Let $\mathcal{Y}=\{\mathcal{Y}_t\}_{t=1}^n$ denote any rainy video in $\mathcal{D}$ or $\mathcal{U}$, where $\mathcal{Y}_t \in \mathcal{R}^{h\times w}$ is the
$t$-th image frame. Similar to~\cite{li2018video,li2016rain}, we decompose the rainy video $\mathcal{Y}$ into three parts, i.e.,
\begin{equation}
    \mathcal{Y} = f(\mathcal{Y};W) + \mathcal{R} + \mathcal{E}, ~ ~ \mathcal{E}_{ijt} \sim \mathcal{N}(0, \sigma^2),
    \label{eq:decompose}
\end{equation}
where $f(\mathcal{Y};W)$, $\mathcal{R}$ and $\mathcal{E}$ are the recovered rain-free background, rain layer and residual term, respectively,
and $\mathcal{E}_{ijt}$ is the element of $\mathcal{E}$ at location $(i,j,t)$.
The residual term is assumed to follow a zero-mean Gaussian distribution with variance $\sigma^2$.
$f(\cdot;W)$, which is parameterized by DNNs, denotes a function that maps the observed rainy video to the underlying rain-free background, and is called the ``derainer'' in
this paper. Next, we consider how to model the derainer parameter $W$ and rain layer $\mathcal{R}$:

\noindent \textbf{Modeling of background layer:} As is well known, one general prior knowledge for video data is that the rain-free background video has strong correlations and similarities
along spatial and temporal dimensions. Therefore, for any rainy video $\mathcal{Y} \in \mathcal{U}$, we encode such a knowledge through the following MRF prior
distribution for $W$:
\begin{small}
\begin{equation}
    p(W) \propto \exp \left(- \rho\sum_{i,j,t}\bm{v}^T\bm{\gamma}\right),
    \label{eq:MRF_prior}
\end{equation}
\end{small}
\hspace{-2mm}where $\bm{v}=\left[\begin{smallmatrix}|f_{i+1,j,t}-f_{ijt}| \\ |f_{i,j+1,t}-f_{ijt}| \\ |f_{i,j,t+1}-f_{ijt}|\end{smallmatrix}\right]$,
$\bm{\gamma}=\left[\begin{smallmatrix} \gamma_1 \\ \gamma_2 \\ \gamma_3\end{smallmatrix}\right]$,
and $f_{ijt}$ denotes the element of $f(\mathcal{Y};W)$ at location $(i, j, t)$.  $\rho$ and $\bm{\gamma}$ are both manual hyper-paramerters, and the latter
represents the strength of smoothness constraint on the spatial and temporal dimensions.
As for the rainy video $\mathcal{Y}\in\mathcal{D}$, the known rain-free background $\mathcal{X}$ can be further
embedded into Eq.~\eqref{eq:MRF_prior} as another strong prior, i.e.,
\begin{small}
\begin{equation}
    p(W) \propto \exp\left( - \frac{\Vert f(\mathcal{Y};W) - \mathcal{X}\Vert_2}{\varepsilon_0^2}- \rho\sum_{i,j,t}\bm{v}^T\bm{\gamma}\right),
    \label{eq:MRF_Gauss_prior}
\end{equation}
\end{small}
\hspace{-2mm} where $\varepsilon_0$ is a very small hyper-paramerter close to zero.

As for the derainer $f(\cdot;W)$, we adopt a simple network architecture as shown in Fig.~\ref{fig:derain_net}. Without any special designs, it only contains 
several 3-D convolution layers and residual blocks~\cite{he2016identity}. To accelerate the computation, the pixel-unshuffle~\cite{zhang2018ffdnet} and
pixel-shuffle~\cite{shi2016real} layers are added to the head and the tail of the network, respectively.

\noindent \textbf{Modeling of rain layer:} Intuitively, the rain layer is a dynamical sequence, thus we naturally employ the spatial-temporal
process~\cite{doretto2003dynamic,xie2019learning,xie2020aaai} in statistics to characterize it. Let's use $\mathcal{R}_t$ to denote the $t$-th image frame of rain layer sequence $\mathcal{R}$,
and then our dynamical rain generator can be formulated as follows,
\begin{align}
    \bm{s}_t &= F(\bm{s}_{t-1}, \bm{z}_t;\bm{\alpha}),  \label{eq:state_generator}\\
    \mathcal{R}_t &= H(\bm{s}_{t};\bm{\beta}),  \label{eq:rain_generator}
\end{align}
where
\begin{equation}
    \bm{z}_t \sim \mathcal{N}(0, \bm{I}), ~ \bm{s}_0 \sim \mathcal{N}(0, \bm{I}),
    \label{eq:state_latent}
\end{equation}
$\bm{s}_t$ represents the hidden state variable in $t$-th frame, and $\bm{z}_t$ the noise vector.
Specifically, Eq.~\eqref{eq:state_generator} is the transition model with parameters $\bm{\alpha}$ expecting to depict the dynamics of rains over time,
and Eq.~\eqref{eq:rain_generator} is the emission model with parameters $\bm{\beta}$ that maps the hidden state space to the space of rain layer.
Note that the noise vectors $\{\bm{z}_t\}_{t=1}^n$ are independent
of each other, and each $z_t$ encodes the random factors that affect the rains (e.g., wind, camera motion, etc) at time $t$ in the transition from $\bm{s}_{t-1}$ to $\bm{s}_t$.

Furthermore, following \cite{xie2019learning}, we can extend the generator to an advanced version for multiple rain videos.
Specifically, for the $i$-th rain video $\mathcal{R}^i=\{\mathcal{R}_t^i\}_{t=1}^n$, another vector $\bm{m}^i \sim \mathcal{N}(0,\bm{I})$ is introduced to
account for the variation of rain appearances or patterns over different videos, and thus the transition model of Eq.~\eqref{eq:state_generator} can be reformulated as:
\begin{equation}
    \bm{s}_t^i = F(\bm{s}_{t-1}^i, \bm{z}_t^i, \bm{m}^i;\bm{\alpha}),  
    \label{eq:state_generator_multi}
\end{equation}
where $\bm{m}^i$ is fixed for the $i$-th rain video. For notation convenience, we write Eqs.~\eqref{eq:state_generator_multi} and \eqref{eq:rain_generator}
together as follows: 
\begin{equation}
    \mathcal{R}^i = G(\bm{s}^i_0, \bm{z}^i, \bm{m}^i;\bm{\theta}),  
    \label{eq:generator_all_simple_format}
\end{equation}
where $\bm{z}^i=\{\bm{z}^i_t\}_{t=1}^n$, $\bm{\theta}=\{\bm{\alpha}, \bm{\beta}\}$. In practice, we use the extended version of Eq.~\eqref{eq:generator_all_simple_format}
to simultaneously fit the rain layers in each mini-batch of video data.

To increase the capacity of such a dynamical rain generator, we parameterize both of the transition model and the emission model by DNNs. Following~\cite{xie2019learning}, we use a two-layers mutli-layer perceptron (MLP) in Fig.~\ref{fig:generator} (a) for the transition model. As to the emission model, we elaborately design a
CNN architecture that takes the state variable $\bm{s}_t$ as input and outputs the rain image as shown in Fig.~\ref{fig:generator} (b),
which is mainly inspired by a recent work~\cite{wang2020rain} that uses CNN as a latent variable model to generate rain streaks.

\noindent\textbf{Remark:} The employment of such a dynamical generator to fit the rain layers is one of the main contributions of this work, which directly affects the deraining
performance of the entire model. Therefore, it is necessary to validate the capability of the dynamical generator on simulating the rain layers. To prove this point, 
we pre-collected some rain layer videos synthesized by commercial Adobe After Effects\footnote{\scriptsize\url{https://www.adobe.com/products/aftereffects.html}} software
from YouTube as source videos, and trained the dynamical generator on them. Empirically, we found that the proposed dynamical rain generator is able to sufficiently mimic the given rain layer videos. Due to the page limitation, we put the experiments to the supplementary materials. 

\subsection{Maximum A Posteriori Estimation}
Combining Eqs.~\eqref{eq:decompose}-\eqref{eq:state_latent}, a full probabilistic model is obtained for
video deraining. Then our goal turns to maximize the posteriors w.r.t the model parameters $W$ and $\bm{\theta}$, i.e.,
\begin{small}
\begin{align}
    \max_{W,\bm{\theta}} \log p(W, \bm{\theta}|\mathcal{Y}) & = \log p(\mathcal{Y}|W,\bm{\theta}) + \log p(W) + \text{const} \notag \\
                                                            & \triangleq \mathcal{L}(\mathcal{Y};W,\bm{\theta}),
    \label{eq:MAP}
\end{align}
\end{small}
\hspace{-1mm}where $p(\mathcal{Y}|W,\bm{\theta})$ is the likelihood of the rainy video $\mathcal{Y}$. According to Eqs.~\eqref{eq:decompose} and \eqref{eq:generator_all_simple_format},
it can be written as:
\begin{shrinkeq}{-1mm}{-3mm}
\begin{small}
\begin{align*}
    p(\mathcal{Y}|W,\bm{\theta}) &=  \int p(\mathcal{Y}|W,\bm{\theta},\bm{z})p(\bm{z}) \dif \bm{z}  \\
                                 &= \int \mathcal{N}\left(f(\mathcal{Y};W)+G(\bm{s}_0,\bm{z};\bm{\theta}), \sigma^2\bm{I}\right)p(\bm{z}) \dif \bm{z}.
\end{align*}
\end{small}
\end{shrinkeq}

Finally, we directly optimize the problem of Eq.~\eqref{eq:MAP} on the whole labeled and unlabeled data sets, i.e.,
\begin{small}
\begin{equation}
    \max_{W,\bm{\theta}} \sum\limits_{\mathcal{Y}^k \in \mathcal{D}} \mathcal{L}(\mathcal{Y}^k;W, \bm{\theta}) + 
                         \sum\limits_{\mathcal{Y}^k \in \mathcal{U}} \mathcal{L}(\mathcal{Y}^k;W, \bm{\theta}).  
    \label{eq:loss_func}
\end{equation}
\end{small}
\hspace{-1mm}The insight behind Eq.~\eqref{eq:loss_func} is to learn a general mapping from rainy videos to clean ones, based on large amount of data
samples in $\mathcal{D}$ and $\mathcal{U}$, which is expected to obtain a more efficient and robust derainer than that in traditional inference paradigm implementing on
single video.

Most notably, if only considering labeled data set, our method naturally degenerates
into a supervised deraining model. However, involving the unlabeled real data can increase the generalization capacity of the model such that it can be applied to the real cases as shown in the ablation studies
in Sec.~\ref{subsubsec:ablation}.

\subsection{Inference and Learning Algorithm}\label{subsec:inference}
For notation brevity, we only consider one data sample $\mathcal{Y}$ in this part.
Inspired by the technology of alternative back-propagation through time~\cite{xie2019learning}, a Monte Carlo-based EM~\cite{dempster1977maximum} algorithm is designed
to maximize $\mathcal{L}(\mathcal{Y};W,\bm{\theta})$,
in which the expectation step samples the latent variable $\bm{z}$ from
the posterior distribution $p(\bm{z}|\mathcal{Y})$, and the maximization step updates the model parameters $W$ and $\bm{\theta}$ based on the inferred latent variable $\bm{z}$.

\vspace{1mm} \noindent \textbf{E-Step:}
Let $(W^{\text{old}}, \bm{\theta}^{\text{old}})$ and $p_{\text{old}}(\bm{z}|\mathcal{Y})$ denote the current model parameters and the posterior under them,
we can sample $\bm{z}$ from $p_{\text{old}}(\bm{z}|\mathcal{Y})$ using the Langevin dynamics~\cite{langevin}:
\begin{small}
\begin{align}
    \bm{z}^{(\tau+1)} &= \bm{z}^{(\tau)} + \frac{\delta^2}{2} \left[\frac{\partial}{\partial \bm{z}} \log p_{\text{old}}(\bm{z}|\mathcal{Y})\right]\bigg\vert_{\bm{z}=\bm{z}^{(\tau)}}
                                         + \delta \bm{\xi}^{(\tau)} \notag \\
                      &= \bm{z}^{(\tau)} - \frac{\delta^2}{2} \left[\frac{\partial}{\partial \bm{z}} g(\bm{z})\right]\bigg\vert_{\bm{z}=\bm{z}^{(\tau)}} + \delta \bm{\xi}^{(\tau)}
    \label{eq:EStep}
\end{align}
\end{small}
\hspace{-1mm}where we define
\begin{small}
\begin{equation}
    g(\bm{z}) = \frac{1}{2\sigma^2}\left\Vert \mathcal{Y}-f(\mathcal{Y};W^{\text{old}})-G(\bm{s}^0,\bm{z};\bm{\theta}^{\text{old}}) \right\Vert_2 + \frac{1}{2}\Vert \bm{z} \Vert_2,
    \label{eq:EStep_gz}
\end{equation}
\end{small}
\hspace{-1mm}$\tau$ indexs the time step for Langevin dynamics, $\delta$ denotes the step size, and $\bm{\xi}^{(\tau)}$ is the Gaussian white noise, which is used to avoid falling into local modes. A key point to compute Eq.~\eqref{eq:EStep} is
$\frac{\partial}{\partial \bm{z}} \log p_{\text{old}}(\bm{z}|\mathcal{Y})=\frac{\partial}{\partial \bm{z}} \log p_{\text{old}}(\mathcal{Y}, \bm{z})$, and the right term can 
be easily calculated. 

\begin{algorithm}[tb]
    \caption{Inference and learning procedure for S2VD}
    \label{alg:EM_algotithm}
    \begin{algorithmic}[1]
        \REQUIRE training data $\mathcal{D}=\{\mathcal{Y}^{b_j}, \mathcal{X}^{b_j}\}_{j=1}^{B_l}$ and $\mathcal{U}=\{\mathcal{Y}^{b_j}\}_{j=B_l+1}^{B_l+B_u}$,
        where $\mathcal{Y}^{b_j}$ denotes the $j$-th mini-batch data, number of Langevin steps $l$. \\
        \ENSURE the derainer parameters $W$. \\
        \STATE Initialize  $W$ and $\bm{\theta}^{b_j}$, $\bm{z}^{b_j}$, $j=1,2,\cdots,B_l+B_u$. \\
        \WHILE {not converged}
            \FOR {$j=1, 2, \cdots, B_l+B_u$}
                \STATE Sample the mini-batch data $\{\mathcal{Y}^{b_j}, \mathcal{X}^{b_j}\}$ or $\mathcal{Y}^{b_j}$. \\
                \STATE \textbf{E-Step:} For each data example $\mathcal{Y}^i$ in current mini-batch $\mathcal{Y}^{b_j}$, run $l$ steps of Langevin dynamics
                to sample $\bm{z}^i$ following Eq.~\eqref{eq:EStep_gz}. \\
                \STATE \textbf{M-Step:} Update $W$ and $\bm{\theta}^{b_j}$ by Eq.~\eqref{eq:MStep_update}. \\
            \ENDFOR
        \ENDWHILE
    \end{algorithmic}
\end{algorithm} 
In practice, for the purpose of avoiding the high computational cost of MCMC, At each learning iteration, Eq.~\eqref{eq:EStep} starts from the previous updated results of $\bm{z}$.
As for the initial state vector $\bm{s}_0$ and the rain variation vector $\bm{m}$ in Eq.~\eqref{eq:generator_all_simple_format}, because they are also latent variables in our model, we sample them together with
$\bm{z}$ using the Langevin dynamics.

\begin{table*}[t]
    \centering
    \caption{PSNR/SSIM results of different methods on the synthetic testing data set of \textit{NTURain}. The best and second best results are highlighted in
            \textcolor{red}{red} and \textcolor{blue}{blue}, respectively.}
    \scriptsize
    \begin{tabular}{@{}C{0.7cm}@{}|@{}C{1.00cm}@{}@{}C{1.00cm}@{}|@{}C{1.05cm}@{}@{}C{1.05cm}@{}|@{}C{1.05cm}@{}@{}C{1.05cm}@{}|@{}C{1.05cm}@{}@{}C{1.05cm}@{}|
                                   @{}C{1.05cm}@{}@{}C{1.05cm}@{}|@{}C{1.05cm}@{}@{}C{1.05cm}@{}|@{}C{1.00cm}@{}@{}C{1.00cm}@{}|@{}C{1.00cm}@{}@{}C{1.00cm}@{}}
        \Xhline{0.8pt}
        \multirow{2}*{\makecell{\textit{Clip} \\ \textit{No.}}}
        & \multicolumn{2}{c|}{Rain}   & \multicolumn{2}{c|}{DSC~\cite{luo2015removing}}     & \multicolumn{2}{c|}{FastDerain~\cite{Jiang2019}}
                                      & \multicolumn{2}{c|}{DDN~\cite{fu2017removing}}      & \multicolumn{2}{c|}{PReNet~\cite{ren2019progressive}}
                                      & \multicolumn{2}{c|}{SpacCNN~\cite{chen2018robust}}  & \multicolumn{2}{c|}{SLDNet~\cite{yang2020self}}
                                      & \multicolumn{2}{c}{S2VD} \\
        \Xcline{2-17}{0.4pt}
                 &PSNR  &SSIM   &PSNR  &SSIM   &PSNR  &SSIM   &PSNR  &SSIM   &PSNR  &SSIM            &PSNR  &SSIM
                 &PSNR                    &SSIM                      &PSNR                    &SSIM \\
        \hline \hline
        a1       &29.71 &0.9149 &27.15 &0.9079 &29.29 &0.9159 &31.79 &0.9481 &32.13 &\textcolor{blue}{0.9511} &30.57 &0.9334 
                 &\textcolor{blue}{33.72} &0.9508          &\textcolor{red}{36.39} &\textcolor{red}{0.9658} \\
        a2       &29.30 &0.9284 &28.84 &0.9224 &30.21 &0.9245 &30.34 &0.9360 &30.41 &0.9375          &31.29 &0.9356
                 &\textcolor{red}{33.82} &\textcolor{blue}{0.9512}   &\textcolor{blue}{33.06} &\textcolor{red}{0.9519} \\
        a3       &29.08 &0.8964 &26.73 &0.8942 &29.94 &0.9039 &30.70 &0.9301 &30.73 &0.9316          &30.63 &0.9247
                 &\textcolor{blue}{33.12} &\textcolor{blue}{0.9404}  &\textcolor{red}{35.75}  &\textcolor{red}{0.9564} \\
        a4       &32.62 &0.9381 &30.58 &0.9381 &34.69 &0.9707 &35.77 &0.9689 &35.77 &0.9700          &35.30 &0.9620
                 &\textcolor{blue}{37.35} &\textcolor{blue}{0.9722}  &\textcolor{red}{39.53}  &\textcolor{red}{0.9779} \\
        \hline
        b1       &30.03 &0.8956 &30.06 &0.9015 &29.35 &0.9139 &32.53 &0.9465 &32.66 &\textcolor{blue}{0.9491} &32.26 &0.9454
                 &\textcolor{blue}{34.21} &0.9482                    &\textcolor{red}{37.34}  &\textcolor{red}{0.9712} \\
        b2       &30.69 &0.8874 &30.85 &0.9017 &31.90 &0.9520 &33.89 &0.9559 &33.74 &0.9557         &35.11 &\textcolor{blue}{0.9677}
                 &\textcolor{blue}{35.80} &0.9595                    &\textcolor{red}{40.55}  &\textcolor{red}{0.9821} \\
                 b3       &32.31 &0.9299 &31.30 &0.9295 &29.28 &0.9287 &35.38 &0.9663 &35.34 &\textcolor{blue}{0.9681} &34.69 &0.9566
                 &\textcolor{blue}{36.34} &0.9614                    &\textcolor{red}{38.82} &\textcolor{red}{0.9754} \\
                 b4       &29.41 &0.8933 &30.61 &0.9089 &27.70 &0.9095 &32.62 &0.9462 &33.17 &0.9526          &\textcolor{blue}{34.87} &\textcolor{blue}{0.9536}
                 &33.85   &0.9469          &\textcolor{red}{37.53} &\textcolor{red}{0.9657} \\
        \hline
        avg.     &30.41 &0.9108 &29.52 &0.9130 &30.54 &0.9255 &32.87 &0.9497 &32.99 &0.9519
                 &33.11 &0.9475 &\textcolor{blue}{34.89} &\textcolor{blue}{0.9540} &\textcolor{red}{37.37} &\textcolor{red}{0.9683} \\
        \Xhline{0.8pt}
    \end{tabular}
    \label{tab:NTU-synthetic}
    \vspace{-3mm}
\end{table*}
\begin{figure*}[t]
    \centering
    \includegraphics[scale=0.89]{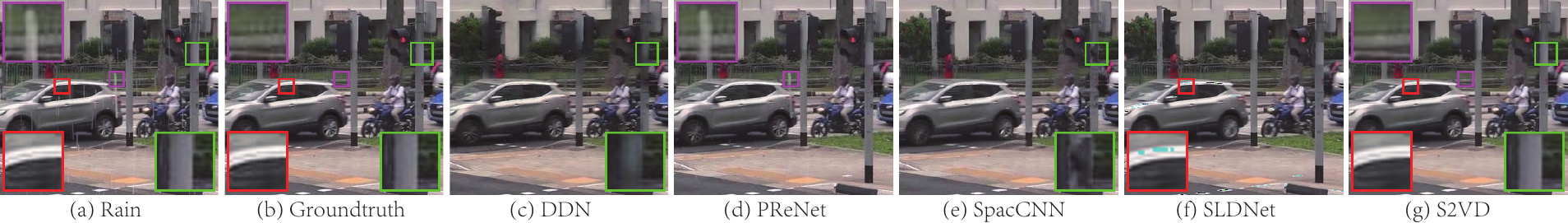}
    \vspace{-1mm}
    \caption{Qualitative results of different methods on one typical image in \textit{NTURain} synthetic testing data set. From left to right: (a) rainy image,
        (b) ground truth image, (c)-(g) deraining results by DDN, PReNet, SpacCNN, SLDNet and our S2VD.}
    \label{fig:ntu_syn_a4}
\end{figure*}

\vspace{1mm} \noindent \textbf{M-Step:} Denote the sampled latent variable in E-Step as $\tilde{\bm{z}}$, M-Step aims to maximize the approximate lower bound w.r.t. $W$ and
$\bm{\theta}$ as follows:
\begin{small}
\begin{align}
    \max_{W, \bm{\theta}} \mathcal{Q}(W,\bm{\theta}) &= \int p_{\text{old}}(\bm{z}|\mathcal{Y}) \log p(\mathcal{Y},\bm{z}|W,\bm{\theta}) \dif \bm{z} + \log p(W)  \notag \\
                                                     &\approx \log p(\mathcal{Y},\tilde{\bm{z}}|W,\bm{\theta}) + \log p(W).
    \label{eq:Q-fun-max}
\end{align}
\end{small}
\hspace{-1mm}Equivalently, Eq.~\eqref{eq:Q-fun-max} can be further rewritten as the following minimization problem, i.e.,
\begin{small}
\begin{align}
    \min_{W, \bm{\theta}} \hat{\mathcal{L}}(W, \bm{\theta}) &= \frac{1}{2\sigma^2}\left\Vert \mathcal{Y}-f(\mathcal{Y};W)-G(\tilde{\bm{z}},\bm{s}^0;\bm{\theta}) \right\Vert_2 + \notag  \\
    &\mathrel{\phantom{=}} \hspace{-1mm}\rho \sum_{i,j,t} \bm{v}^T \bm{\gamma} 
    + 1_{[\mathcal{Y}\in \mathcal{D}]} \cdot \frac{\Vert f(\mathcal{Y};W) - \mathcal{X}\Vert_2}{\varepsilon_0^2},
    \label{eq:MStep}
\end{align}
\end{small}
\hspace{-1mm}where $1_{[\mathcal{Y}\in \mathcal{D}]}$ equals to 1 when $\mathcal{Y}$ comes from the labeled data set $\mathcal{D}$ otherwise 0. 
Naturally, we can update $W$ and $\bm{\theta}$ by gradient descent based on the back-propagation (BP) algorithm~\cite{rumelhart1986learning} as follows,
\begin{small}
\begin{equation}
    \varLambda \leftarrow \varLambda - \eta \frac{\partial}{\partial \varLambda} \hat{\mathcal{L}}(W, \bm{\theta}), ~ \varLambda \in \{W, \bm{\theta}\},
    \label{eq:MStep_update}
\end{equation}
\end{small}
where $\eta$ denotes the step size.

Due to the capacity limitation, we empirically find it difficult to fit
the rain layers in all training videos using only one single generator defined in Eq.~\eqref{eq:generator_all_simple_format}. Therefore, we train one
generator for each mini-batch data. With such a strategy, our model performs well  throughout all our experiments. The mini-batch size is 12.
A detailed description of the proposed algorithm is presented in Algorithm~\ref{alg:EM_algotithm}.
\section{Experimental Results} \vspace{-2mm}
In this section, we conduct some experiments to evaluate the effectiveness of the proposed semi-supervised
video deraining model on synthetic and real data sets. 
And we briefly denote our \textbf{S}emi-\textbf{S}upervised \textbf{V}ideo \textbf{D}eraining model by S2VD in the following presentation.

\begin{figure*}[t]
    \centering
    \includegraphics[scale=0.89]{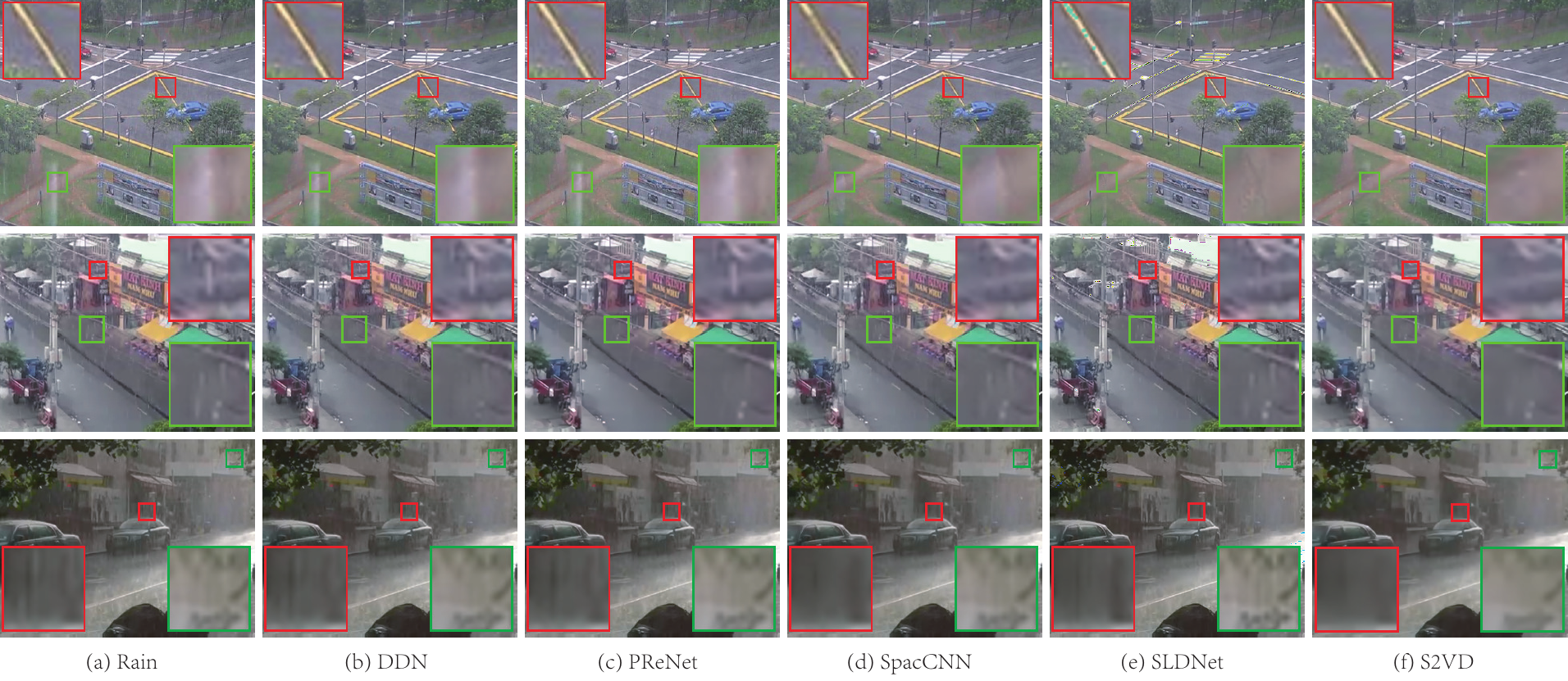}
    \vspace{-2mm}
    \caption{Visual comparisons of different methods on three typical real testing images from \textit{NTURain} \cite{chen2018robust} (the 1st row) and
    \cite{li2018video} (the 2nd and 3rd row). From left to right: (a) rainy image, (b)-(f) deraining results by DDN, PReNet, SpacCNN, SLDNet and our S2VD.}
    \label{fig:ntu_real_ra4}
    \vspace{-3mm}
\end{figure*}
\begin{figure*}[t]
    \centering
    \includegraphics[scale=0.68]{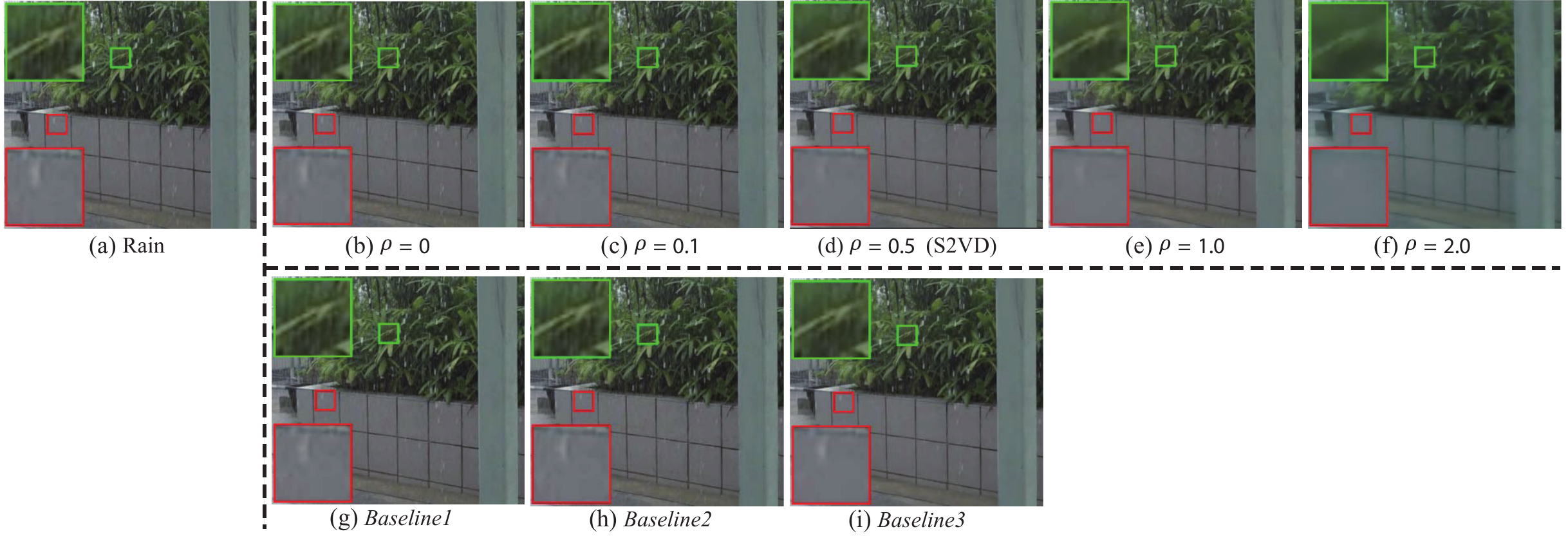}
    \vspace{-1mm}
    \caption{Comparisons of S2VD under different settings: (a) rainy image, (b)-(f) deraining
    results of S2VD with different $\rho$ values, (g)-(i) deraining results of different \textit{Baseline}s defined in Sec.~\ref{subsubsec:ablation}.} 
    \label{fig:rho_hyper}
    \vspace{-3mm}
\end{figure*}

\subsection{Evaluation on Rain Removal Task}
\noindent\textbf{Training Details:} To train S2VD, we employ the synthesized training data of \textit{NTURain}~\cite{chen2018robust} as labeled data set, which
contains 8 rain-free video clips of various scenes. For each rain-free video,  3 or 4 rain layers are synthesized by the Adobe
After Effects with different settings,
and then added to the videos as rainy ones. As for unlabeled
data, 7 real rainy videos without ground truths in the testing data of \textit{NTURain} are employed.
To relieve the burden of GPU memory, we use truncated back-propagation through time
in training, meaning that the whole training sequence is divided into different non-overlapped chunks for forward and
backward propagation. The length of each chunk is 20.

The Adam~\cite{Kingma2015} algorithm is used to optimize the model parameters in the M-Step of our algorithm. All the network parameters are initialized by~\cite{Saxe2014}. The initialized learning rates for the transition model, emission model and the derainer are set to be $1e\text{-}3$, $1e\text{-}4$ and $2e\text{-}4$, respectively, and decayed by half after 30 epochs. The mini-batch size is set as 12, and each video is clipped into small
blocks with spatial size $64\times 64$ pixels. Note that we only update the parameter $W$ for the first 5 epochs to pretrain the derainer, which makes the training more stable. As for the hyper-parameters, $\varepsilon_0^2=1e\text{-}6$, $\rho=0.5$, $\bm{\gamma}=[1,1,2]^T$, and more analysis experiments on them are presented in Sec.~\ref{subsec:ablation}.

\vspace{-3mm} \subsubsection{Evaluation on Synthetic Data}\vspace{-1mm}
We test our S2VD on the synthetic testing data set of \textit{NTURain}~\cite{chen2018robust}, which consists of two groups of data sets. The videos in the first
group (with prefix ``a'' in Table~\ref{tab:NTU-synthetic}) are captured by a panning and unstable camera, while those in the second group (with prefix ``b'' in Table~\ref{tab:NTU-synthetic})
by a fast moving camera with a speed range from 20 to 30 \textit{km/h}. As to the methods for comparison, six SOTAs are considered, including one model-based image deraining method DSC~\cite{luo2015removing}, one model-based video deraining method
FastDerain~\cite{Jiang2019}, two DL-based image deraining methods DDN~\cite{fu2017removing} and PReNet~\cite{ren2019progressive}, two DL-based video deraining methods 
SpacCNN~\cite{chen2018robust} and SLDNet~\cite{yang2020self}. The average PSNR and SSIM~\cite{wang2004image} are used as quantitative metrics, which
are evaluated only in the luminance channel since we are sensitive to the luminance information.
\begin{table}[t]
    \centering
    \caption{Average PSNR/SSIM results of S2VD on the synthetic testing dataset of \textit{NTURain} under different $\rho$ values.}
    \label{tab:hyper-tau}
    \scriptsize
    \begin{tabular}{@{}C{1.5cm}@{}|@{}C{1.3cm}@{}|@{}C{1.4cm}@{}|@{}C{1.4cm}@{}|@{}C{1.3cm}@{}|@{}C{1.3cm}@{}}
        \Xhline{0.8pt}
        \multirow{2}*{Metrics} & \multicolumn{5}{c}{$\rho$} \\
        \Xcline{2-6}{0.4pt}
                & 0             & 0.1             & 0.5         & 1        & 2 \\
        \Xhline{0.4pt}                                                     
        PSNR    & 38.18         & 38.05           & 37.37       & 35.50    & 31.55 \\
        \Xhline{0.4pt}                                                     
        SSIM    & 0.9719        & 0.9713          & 0.9683      & 0.9519   & 0.8947  \\
        \Xhline{0.8pt}
    \end{tabular}
    \vspace{-4mm}
\end{table}

Table~\ref{tab:NTU-synthetic} lists the average PSNR/SSIM results on 8 testing videos. Evidently, S2VD attains the best or at least second best performance in all cases. Comparing with current SOTAs (SpacCNN or SLDNet), our method achieves at least 2.5dB PSNR and 0.01 SSIM gains.
The qualitative results are shown in Fig.~\ref{fig:ntu_syn_a4}.
Note that we only display the results of DL-based methods due to the page limitation.
We observe that: 1) The derained results of PReNet still contain some rain streaks. 2) Both DDN and SpacCNN
lose some image contents. 3) SLDNet can not preserve the original color maps very well. However, our S2VD evidently alleviate these deficiencies and obtains the closest results to the ground truths, which verifies the effectiveness of our model.

\begin{table}[t]
    \centering
    \caption{Average PSNR/SSIM results of three baselines and S2VD on the synthetic testing dataset of \textit{NTURain}.}
    \label{tab:ablation}
    \scriptsize
    \begin{tabular}{@{}C{1.5cm}@{}|@{}C{1.8cm}@{}|@{}C{1.8cm}@{}|@{}C{1.8cm}@{}|@{}C{1.4cm}@{}}
        \Xhline{0.8pt}
        \multirow{2}*{Metrics} & \multicolumn{4}{c}{Methods} \\
        \Xcline{2-5}{0.4pt}
                & \textit{Baseline1} & \textit{Baseline2}     & \textit{Baseline3}   & S2VD \\
        \Xhline{0.4pt}
        PSNR    &      36.11         &        37.12           &        37.96         & 37.37    \\
        \Xhline{0.4pt}
        SSIM    &      0.9602        &        0.9673          &        0.9717        & 0.9683    \\
        \Xhline{0.8pt}
    \end{tabular}
    \vspace{-4mm}
\end{table}

\vspace{-3mm} \subsubsection{Evaluation on Real Data}\vspace{-1mm}
To further test the generalization capability of S2VD in real tasks, we evaluate it on two kinds of real rainy videos, i.e., the real testing data set in \textit{NTURain} and several other
real rainy videos in ~\cite{li2018video}. Note that the former is included in our training set as unlabeled data, but the latter is not.
Fig.~\ref{fig:ntu_real_ra4} presents some typical deraining results by different methods on these two kinds of 
data sets. It can be seen that S2VD achieves the best visual results comparing with other methods.
Especially, the superiority of our model shown in the second data set substantiates that S2VD is able to handle the real rainy videos even though they do not appear in the unlabeled data set. This generalization capability would be potentially useful in real deraining tasks.

\subsection{Additional Analysis}\label{subsec:ablation} 
\subsubsection{Sensitiveness of Hyper-paramerter $\rho$} \label{subsubsec:hyper-rho} 
The hyper-paramerter $\rho$ in Eq.~\eqref{eq:MRF_prior} or \eqref{eq:MRF_Gauss_prior} controls the relative importantance of MRF prior in S2VD.
The quantitative performances on the synthetic testing data set and the qualitative performances on the real testing data set of \textit{NTURain} under different $\rho$ values are
presented in Table~\ref{tab:hyper-tau} and Fig.~\ref{fig:rho_hyper}, respectively. On one hand, when $\rho$ increases, the performance on the synthetic testing set tends
to decrease as shown in Table~\ref{tab:hyper-tau}, since the relative importance of the constraint built on the ground truth in Eq.~\eqref{eq:MRF_Gauss_prior} decreases. On the other hand, 
the MRF prior is able to prevent the derainer from overfitting the synthetic data, thus improving the generalization capability in real cases, which has been sufficiently
verified by the visual comparisons in Fig.~\ref{fig:rho_hyper}. By taking into account these two aspects, we set $\rho$ as $0.5$.

\vspace{-1mm}\subsubsection{Ablation Studies} \label{subsubsec:ablation}
As shown in Eq.~\eqref{eq:MStep}, our S2VD degenerates into the Mean Squre Error (MSE) loss when $\varepsilon_0 \rightarrow 0$.
Comparing with such special case, our model introduces one more likelihood term, one more MRF regularizer and 
the semi-supervised learning paradigm. To clarify the effect of each part, we compare S2VD with three baselines as follows:
1) \textit{Baseline1:} We only train the derainer with the MSE loss on labeled data set.
2) \textit{Baseline2:} We train S2VD with $\varepsilon_0^2=1e\textit{-}6$ and $\rho=0$ only on labeled data set
so that we can justify the marginal gain from the likelihood term by comparing with \textit{Baseline1} using the MSE loss.
3) \textit{Baseline3:} On the basis of \textit{Baseline2}, we introduce the MRF regularizer with $\rho=0.5$. 

The quantitative comparisons on synthetic testing data set of \textit{NTURain} are listed in Table~\ref{tab:ablation}, and the visual results on
real testing data set are also displayed in Fig.~\ref{fig:rho_hyper}. In summary, we can see that:
1) The performance improvement (1.01dB PSNR and 0.0071 SSIM)
of \textit{Baseline2} beyond \textit{Baseline1} substantiates that the likelihood term plays an important role in our model.
2) Under the supervised learning manner, the MRF prior is beneficial to our model in both synthetic and real cases according to the performance of \textit{Baseline3}.
3) The addition of unlabeled data in S2VD increases the generalization capability in real tasks as shown in Fig.~\ref{fig:rho_hyper} (d) and (i).
However, it leads to a little deterioration of the performance on synthetic data, mainly because there is a gap between the rain types contained in the labeled synthetic 
and unlabeled real data sets.

\begin{figure}[t]
    \centering
    \includegraphics[scale=0.66]{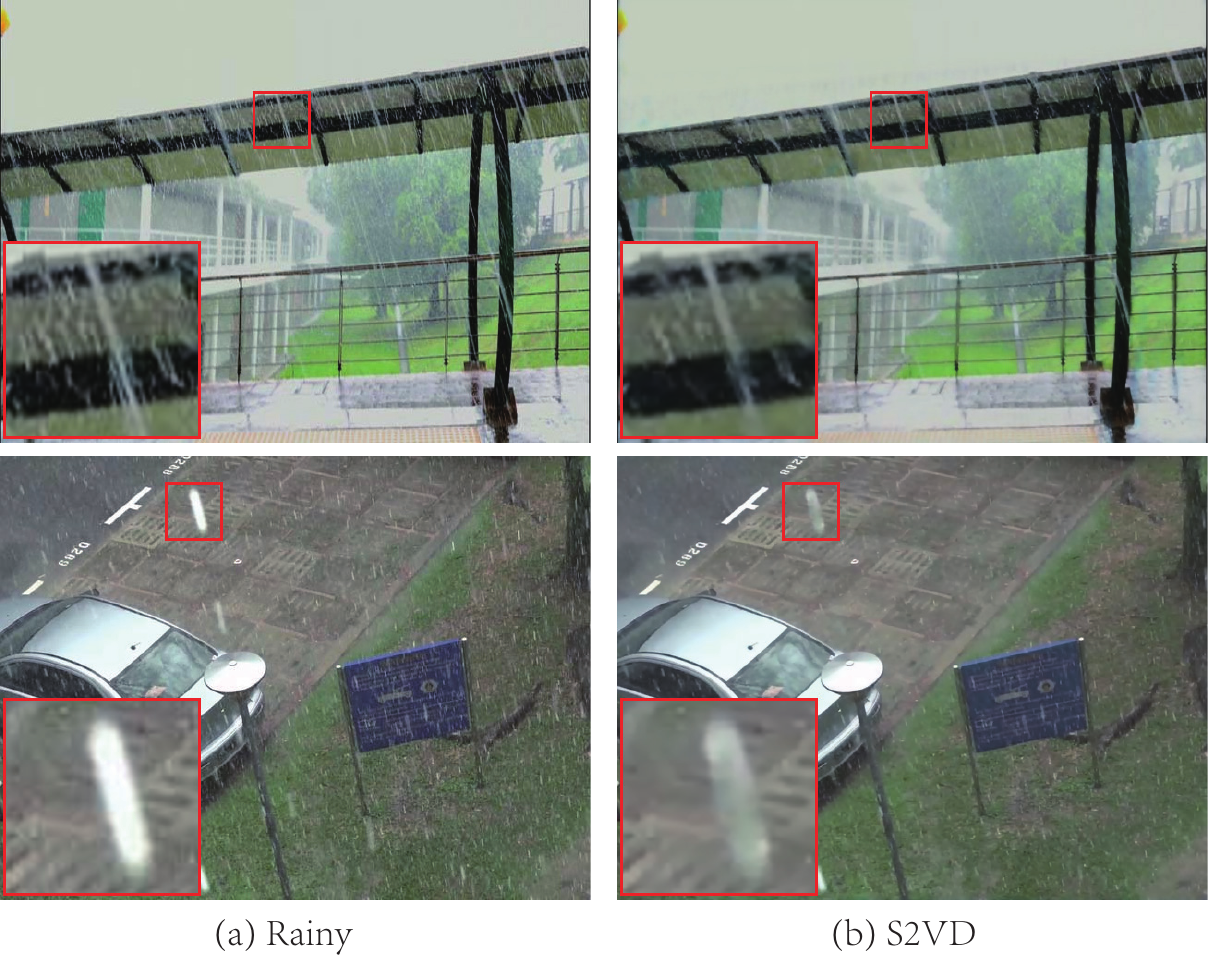}
    \caption{Two typical failure deraining examples by our method. The first row shows a case of large camera motion, while the second row shows another case with heavy rain streaks.}
    \label{fig:limit}
\end{figure}

\vspace{-2mm}\subsubsection{Limitation and Future Direction} \label{subsubsec:limitation} 
Although our method can achieve impressive deraining results as shown above, it may still fails in some real scenarios, e.g., a case with a large camera motion
between adjacent time frames and a case with heavy rain streaks. Two failure examples are shown in Fig.~\ref{fig:limit}. The reason is because the adopted MRF prior for unlabeled real data
is not strong enough to guarantee satisfactory deraining results in these complicated cases. Therefore, it is urgent and necessary to explore better
prior knowledge in order to handle more complicated real deraining tasks in the future. 

\section{Conclusion} 
In this paper, we design a dynamical rain generator based on the spatial-temporal process in statistics. With such a dynamical generator, a semi-supervised
video deraining method is proposed. Specifically, we represent the sequence of rain layers in rain videos using the dynamical rain generator, which is able to facilitate the rain removal task. To handle the generalization issue for real cases, we propose a semi-supervise learning manner
to exploit the common knowledge underlying the synthetic labeled and real unlabeled data sets. Besides, a Monte Carlo-based EM algorithm is designed to learn the model parameters. Extensive experimental results demonstrate the effectiveness of 
the proposed video deraining method.

\noindent \textbf{Acknowledgement:} This research was supported by the National Key R\&D Program of China (2020YFA0713900), the China NSFC projects
under contracts 11690011, 61721002, U1811461, 62076196.

{\small
\bibliographystyle{ieee_fullname}
\bibliography{S2VD}
}

\end{document}